\documentclass[letterpaper]{article} % DO NOT CHANGE THIS
\usepackage{aaai24}  % DO NOT CHANGE THIS
\usepackage{times}  % DO NOT CHANGE THIS
\usepackage{helvet}  % DO NOT CHANGE THIS
\usepackage{courier}  % DO NOT CHANGE THIS
\usepackage[hyphens]{url}  % DO NOT CHANGE THIS
\usepackage{graphicx} % DO NOT CHANGE THIS
\usepackage{natbib}  % DO NOT CHANGE THIS AND DO NOT ADD ANY OPTIONS TO IT
\usepackage{caption} % DO NOT CHANGE THIS AND DO NOT ADD ANY OPTIONS TO IT
\usepackage{algorithm}
\usepackage{algorithmic}
\usepackage{amsmath}
\usepackage{amssymb}
\usepackage{pdfpages}
\usepackage{graphicx}
\usepackage{svg}
\usepackage{cleveref}
\usepackage{makecell}
\usepackage{dsfont}
\usepackage{booktabs}
\usepackage{color,xcolor}

\usepackage{titlesec}
\titlespacing*{\section}{0pt}{*0.75}{*0.75}

\usepackage{newfloat}
\usepackage{listings}
\DeclareCaptionStyle{ruled}{labelfont=normalfont,labelsep=colon,strut=off} % DO NOT CHANGE THIS
\floatstyle{ruled}
\newfloat{listing}{tb}{lst}{}
\floatname{listing}{Listing}
\usepackage[font=small,skip=4pt]{caption}
\title{Large Scale Constrained Clustering With Reinforcement Learning}
\author{
    Benedikt Schesch,\textsuperscript{\rm 1,2}
    Marco Caserta\textsuperscript{\rm 1}
}
\affiliations{
    \textsuperscript{\rm 1}Amazon.com\\
    \textsuperscript{\rm 2}ETH Zürich \\
    \{scheschb,mcaserta\}@amazon.lu \\
}

\usepackage{bibentry}
\begin{document}

\maketitle

\begin{abstract}
Given a network, allocating resources at clusters level, rather than at each node, enhances efficiency in resource allocation and usage. In this paper, we study the problem of finding fully connected disjoint clusters to minimize the intra-cluster distances and maximize the number of nodes assigned to the clusters, while also ensuring that no two nodes within a cluster exceed a threshold distance. While the problem can easily be formulated using a binary linear model, traditional combinatorial optimization solvers struggle when dealing with large-scale instances. We propose an approach to solve this constrained clustering problem via reinforcement learning. Our method involves training an agent to generate both feasible and (near) optimal solutions. The agent learns problem-specific heuristics, tailored to the instances encountered in this task. In the results section, we show that our algorithm finds near optimal solutions, even for large scale instances.
\end{abstract}

\vspace{-5mm}
\section{Introduction}

\noindent In Amazon's operations, resources are traditionally allocated on a site-by-site basis. Shifting to a cluster-based model enhances resource distribution efficiency, such as assigning a single technician with rare skills to multiple sites, reducing travel distances, and capping travel times within clusters. This requires the development of an approach that identifies the best clusters in the network, while respecting unique business constraints. 

Mixed-integer solvers, traditionally used for these types of constrained clustering problems, struggle with large-scale instances, due to the NP-hard nature of these problems. Recent trends \cite{Bengio,BelloRL} involve integrating machine learning, particularly reinforcement learning (RL), into a combinatorial optimization problem. Our RL-based approach trains agents to develop specific heuristics, enabling them to handle large-scale instances effectively.

In the sequel, we present a binary linear formulation for the problem; next, we provide a description of our method; a computational section showcases the effectiveness of the method and, finally, we conclude with some remarks.

\section{Problem Formulation}

Our formulation is loosely based on the modularity maximization problem.~\cite{Agarwal2008} Given a set of $n$ sites and their inter-site travel times $d_{ij}$, we minimize the total intra-cluster travel times, while penalizing unclustered sites. In addition, the maximal distance between any two sites assigned to the same cluster must be below a threshold $D$. The decision variable $x_{ij}$ takes value $0$ if sites $i$ and $j$ are assigned to the same cluster, and $1$ otherwise.
\begin{equation*}
\begin{aligned}
&\min \quad \sum_{i=1}^{n}\sum_{j=i}^{n}\left((1-x_{ij})d_{ij}+x_{ij}D)\right)\mathds{1}_{d_{ij} < D}\\
&\textrm{s.t.} \quad 
\begin{aligned}[t]
& x_{ij} = 1 \text{ if } d_{ij} \geq D & i,j \in 1 \ldots n\\
& x_{ij} \leq x_{il} + x_{lj} & i,j,l \in 1 \ldots n\\
& x_{ij} = x_{ji} & i,j \in 1 \ldots n
\end{aligned}
\end{aligned}
\end{equation*}
In the model, the objective function only considers pair of sites whose distance is below the threshold $D$, and accounts for the total intra-cluster distances and the penalty for the unclustered sites. The triangle inequality ensures that each cluster is a fully connected graph, while the symmetry constraints reflect the undirected nature of the clusters.

\section{Solution methodology}

\paragraph{RL Environment.} We detail the RL formulation:

\textit{State Space.} Each state is a triplet, including the distance matrix, the edges in the current solution, and the remaining available edges.

\textit{Action Space.} The agent selects from the available edges.

\textit{Transition Function.} After choosing an edge, the edge joins the current solution and is removed from the set of available edges. Further edges needed to satisfy the triangle inequality are added, while edges violating this inequality are removed from the available set.

\textit{Reward.} The reward for an action is the difference 
in the objective function resulting from the transition.

\begin{figure}[htbp]
  \centering
  \includegraphics[width=\columnwidth]{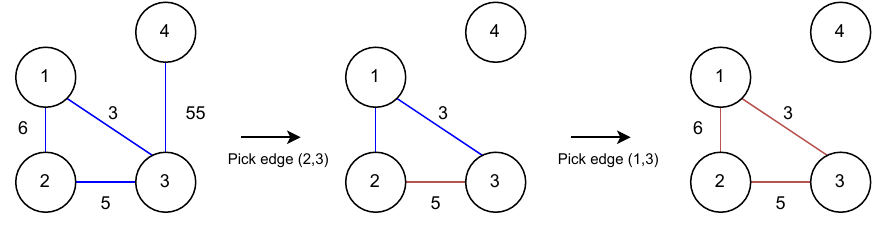}
  \caption{Example instance with threshold $D=60$.}
  \label{example}
\end{figure}

Figure~\ref{example} depicts an example with 4 cities. Blue and red edges are the available edged and the edges in the current solution, respectively. Initially, the agent selects edge (2, 3). Consequently, edge (3, 4) is removed to respect the triangle inequality. In the second step, the agent picks the edge (1, 3) which is added to the solution.  The edge (1, 2) is also added to respect the triangle inequality. Since there are no available edges anymore, a solution has been constructed, with objective function equal to 3 + 5 + 6 + 60 = 74, which is an optimal solution.

This approach allows for easy constraints integration by altering the transition function. It involves removing edges that breach the constraints, with the advantage of not requiring formal constraints expression, which can be complex.

\textbf{Agent.} Our agent, designed to predict the value function and the next edge selection, leverages a graph neural network (GNN) due to the graph-centric nature of our problem. GNNs, recognized for their efficacy in handling graph-structured data \cite{GNN}, are well-suited here. In our model, nodes represent sites and edges denote either available or solution-participating connections. Each edge is characterized by the internode distance and two binary indicators signifying its availability and inclusion in the solution. Each node has a constant feature as a bias term.

The initial feature embedding employs a learned linear transformation. The network architecture has three EGAT convolutions layers \cite{EGAT}. Action readout uses a linear transformation of the final embedding for each available edge, with all embeddings sized at 8. This process assigns scores to edges, guiding the agent's preference. For the critic, we use a dense neural network, processing the flattened distance and availability matrices.

% \begin{figure}[htbp]
% \centering
% \includesvg[width=0.5\textwidth]{agent.svg}
% \caption{Architecture overview}
% \label{agent}
% \end{figure}

\textbf{Training.} We train our agent in our environment using Proximal Policy Optimization (PPO) \cite{PPO} with a learning rate of $10^{-5}$.
% \vspace{-3mm}
\section{Computational Experiments}
We test our approach on two experimental setups, using different distributions over the distance matrices, to verify that  the agent learns distribution specific heuristics. 

\textbf{Cities Environment.}
To mimic the natural distribution of cities, we allocate sites among three major cities on a 2D map ($[0,240]\times[0,240]$). Half of the site are close to the first city, a third to the second, and a sixth to the smallest city. Site locations are normally distributed around city centers, with variances from $\mathcal{U}(80,160)$.

We trained and tested our agent on 128 instances, comparing its performance to that of a random agent. The results for 18-site and 60-site instances are shown in Tables \ref{multimodal 18 and 60}. Our agent consistently outperformed the random agent, achieving a median optimality gap of $0\%$ and $1.02\%$ in the 18-site and the 60-site instances, respectively, with a substantial improvement in the quality of clustering solutions.

% \begin{figure}[H]
%   \centering
%   \includegraphics[width=\columnwidth]{img.pdf}
%   \caption{Example instance with 18 nodes}
%   \label{example multimodal}
% \end{figure}

\begin{table}
\begin{small}
\begin{tabular}{lcccc} \toprule
& \multicolumn{2}{c}{18 sites} & \multicolumn{2}{c}{60 sites}  \\ \cmidrule(r){2-3} \cmidrule(r){4-5}
  & \makecell{Random} & \makecell{Ours} & \makecell{Random} & \makecell{Ours} \\ [0.5ex]
  \midrule
 % \# Sites & 18 & 18 & 60 & 60\\ 
 % \hline
 Mean opt. gap & $16.7\%$ & $\textbf{0.02\%}$ & $31.6\%$ & $\textbf{1.02\%}$ \\
 % \hline
 Median opt. gap & $15.8\%$ & $\textbf{0.00\%}$ & $31.0\%$ & $\textbf{0.43\%}$ \\
 % \hline
 Min opt. gap & $0.00\%$ & $0.00\%$ & $14.3\%$ & $\textbf{0.00\%}$\\
 % \hline
 Max opt. gap & $50.2\%$ & $\textbf{1.45\%}$ & $59.0\%$ & $\textbf{5.11\%}$\\
 \bottomrule
\end{tabular}
\end{small}
\caption{Optimality gaps over 128 instances for the city-like design.}
\label{multimodal 18 and 60}
\end{table}

% \begin{table}[H]
% \begin{tabular}{||c|c|c||} 
%  \hline
%   & Random Agent & Our Agent \\ [0.5ex]
%  \hline\hline
%  Number of Sites & 60 & 60 \\ 
%  \hline
%  Evaluated Instances & 128 & 128 \\
%  \hline
%  Mean Optimality Gap & $31.57\%$ & $\textbf{1.02\%}$ \\
%  \hline
%  Median Optimality Gap & $31.07\%$ & $\textbf{0.43\%}$ \\
%  \hline
%  Minimum Optimality Gap & $14.31\%$ & $\textbf{0\%}$\\
%  \hline
%  Maximum Optimality Gap & $59.01\%$ & $\textbf{5.11\%}$\\
%  \hline
% \end{tabular}
% \caption{Evaluation results for 60 sites}
% \label{multimodal 60}
% \end{table}

\textbf{General Environment.}
In this scenario, the distance matrix is generated with entries following a $\mathcal{U}(0,240)$ distribution. This is a more challenging problem than the cities environment due to its higher dimensionality. We focus on the most difficult instances, for which the Xpress Solver requires over 2000 seconds to terminate with an optimal solution, comprising a dataset of 69 examples with an average runtime of 2276 seconds. We also created a second model, with embedding size equal to 64. 

\begin{table}
\centering
\begin{tabular}{lccc} \toprule
  & \makecell{Random} & \makecell{Ours-8} & \makecell{Ours-64}\\ [0.5ex]
 \midrule
 Mean opt. gap & $5.3\%$ & $3.6\%$ & $\mathbf{3.0}\%$ \\
 Median opt gap & $5.3\%$ & $3.6\%$ & $\mathbf{3.1}\%$ \\
 Min opt gap & $3.9\%$ & $2.7\%$ & $\mathbf{1.8}\%$\\
 Max opt gap & $6.9\%$ & $4.5\%$  & $\mathbf{4.2}\%$\\
 \bottomrule
\end{tabular}
\caption{Optimality gaps on 69 ``difficult'' instances with 64 sites.} 
\label{general 64}
\end{table}

\Cref{general 64} provides the results. Our agent consistently outperforms the random agent in terms of optimality gap, though it does not always reach optimal solutions. The larger agent version shows some improvements, indicating that scaling and further sophistication in the model are beneficial. Despite not achieving optimal results, our agent provides a fast approximation, especially in comparison to the running time of an off-the-shelf solver. Advancements in the graph neural network structure, beyond current EGAT convolutions, could potentially enhance its effectiveness and help closing the gap to optimality.

\vspace{-2mm}
\section{Conclusion}

In this study, we proposed a reinforcement learning-based approach to solve constrained clustering problems, especially useful to deal with very large instances. The results demonstrate that this approach provides near optimal solutions to the constrained clustering problem, with a running time orders of magnitude below the time required by a solver. The next step in our research is to improve the network architecture, to further reduce the optimality gap for difficult instances.

\vspace{-1mm}
\bibliography{aaai24}
\end{document}